\documentclass[runningheads]{llncs}
\usepackage{graphicx}
\usepackage{amsmath,amssymb} 
\usepackage{color}
\usepackage{makecell}
\usepackage{array,multirow,booktabs}
\usepackage{subcaption}
\usepackage{caption}
\graphicspath{{./pics/}}
\usepackage[outdir=./pics/]{epstopdf}
\usepackage{pifont}
\usepackage{hyperref}
\hypersetup{
    colorlinks=true,
    urlcolor=magenta,
}

\newcommand{\dataset}{YouTube-VOS}

\usepackage{xspace}
\makeatletter
\DeclareRobustCommand\onedot{\futurelet\@let@token\@onedot}
\def\@onedot{\ifx\@let@token.\else.\null\fi\xspace}

\def\eg{\emph{e.g}\onedot} 
\def\ie{\emph{i.e}\onedot}

\def\etal{\emph{et al}\onedot}
\makeatother

\begin{document}


\title{~\dataset: A Large-Scale Video Object Segmentation Benchmark} 

\titlerunning{~\dataset}

\author{Ning Xu\inst{1} \and
Linjie Yang\inst{2} \and
Yuchen Fan\inst{3}\and
Dingcheng Yue\inst{3} \and
Yuchen Liang\inst{3} \and
Jianchao Yang\inst{2}\and
Thomas Huang\inst{3}}

\authorrunning{N. Xu~\etal}



\institute{Adobe Research, USA \\
\email{{nxu}}@adobe.com \and
Snapchat Research, USA \\
\email{\{linjie.yang,jianchao.yang\}@snap.com}\\ \and
University of Illinois at Urbana-Champaign, USA\\
\email{\{yuchenf4,dyue2,yliang35,t-huang1\}@illinois.edu}}

\maketitle

\begin{abstract}
Learning long-term spatial-temporal features are critical for many video analysis tasks. However, existing video segmentation methods predominantly rely on static image segmentation techniques, and methods capturing temporal dependency for segmentation have to depend on pretrained optical flow models, leading to suboptimal solutions for the problem. End-to-end sequential learning to explore spatial-temporal features for video segmentation is largely limited by the scale of available video segmentation datasets, i.e., even the largest video segmentation dataset only contains 90 short video clips. To solve this problem, we build a new large-scale video object segmentation dataset called YouTube Video Object Segmentation dataset (\dataset). Our dataset contains 4,453 YouTube video clips and 94 object categories. This is by far the largest video object segmentation dataset to our knowledge and has been released at~\href{http://youtube-vos.org}{http://youtube-vos.org}. We further evaluate several existing state-of-the-art video object segmentation algorithms on this dataset which aims to establish baselines for the development of new algorithms in the future.

\end{abstract}

\section{Introduction}

Learning effective spatial-temporal features has been demonstrated to be very important for many video analysis tasks. For example, Donahue \etal~\cite{donahue2015rcn} propose long-term recurrent convolution network for activity recognition and video captioning. Srivastava \etal\cite{srivastava2015unsupervisedlstm} propose unsupervised learning of video representation with a LSTM autoencoder. Tran \etal~\cite{tran2015c3d} develop a 3D convolutional network to extract spatial and temporal information jointly from a video. Other works include learning spatial-temporal information for precipitation prediction~\cite{xingjian2015convolutional}, physical interaction~\cite{finn2016videopred}, and autonomous driving~\cite{xu2016driving}. 

Video segmentation plays an important role in video understanding, which fosters many applications, such as accurate object segmentation and tracking, interactive video editing and augmented reality. Video object segmentation, which targets at segmenting a particular object instance throughout the entire video sequence given only the object mask on the first frame, has attracted much attention from the vision community recently~\cite{Caelles2017osvos,Perazzi2017masktrack,Yang2018osmn,Cheng2017segflow,Jain_2017_CVPR,Jampani2017vpn,Tokmakov2017memory,Hu2017Maskrnn}. However, existing state-of-the-art video object segmentation approaches primarily rely on single image segmentation frameworks~\cite{Caelles2017osvos,Perazzi2017masktrack,Yang2018osmn}. For example, Caelles \etal~\cite{Caelles2017osvos} propose to train an object segmentation network on static images and then fine-tune the model on the first frame of a test video over hundreds of iterations, so that it remembers the object appearance. The fine-tuned model is then applied to all following individual frames to segment the object without using any temporal information. Even though simple, such an online learning or one-shot learning scheme achieves top performance on video object segmentation benchmarks~\cite{Perazzi2016davis,Jain2014youtubeobjects}. Although some recent approaches~\cite{Jain_2017_CVPR,Cheng2017segflow,Tokmakov2017memory} have been proposed to leverage temporal consistency, they depend on models pretrained on other tasks such as optical flow~\cite{Ilg2017flownet,Revaud2015epicflow} or motion segmentation~\cite{Tokmakov2017mpnet}, to extract temporal information. These pretrained models are learned from separate tasks, and therefore are suboptimal for the video segmentation problem.

Learning long-term spatial-temporal features directly for video object segmentation task is, however, largely limited by the scale of existing video object segmentation datasets. For example, the popular benchmark dataset DAVIS~\cite{Pont-Tuset2017davis} has only 90 short video clips, which is barely sufficient to learn a sequence-to-sequence network from scratch like other video analysis tasks. Even if we combine all the videos from available datasets~\cite{Jain2014youtubeobjects,jumpcut,Fli2013segtrack,brox2010BMS,ochs2014FBMS,galasso2013VSB100}, its scale is still far smaller than many other video analysis datasets such as YouTube-8M~\cite{abu2016youtube} and ActivityNet~\cite{heilbron2015activitynet}. To solve this problem, we present the first large-scale video object segmentation dataset called~\dataset~(YouTube Video Object Segmentation dataset) in this work. Our dataset contains 4,453 YouTube video clips featuring 94 categories covering humans, common animals, vehicles, and accessories. Each video clip is about 3$\sim$6 seconds long and often contains multiple objects, which are manually segmented by professional annotators. Compared to existing datasets, our dataset contains a lot more videos, object categories, object instances and annotations, and a much longer duration of total annotated videos. Table~\ref{tab:dataset-cmp} provides quantitative scale comparisons of our new dataset against existing datasets. The dataset has been released at \href{https://youtube-vos.org}{https://youtube-vos.org}. We elaborate the collection process of our dataset in Section~\ref{sec:dataset}.

In this report, we also retrain state-of-the-art video object segmentation algorithms on~\dataset~and benchmark their performance on the validation set which contains 474 videos. In addition, our validation set contains 26 unique categories that do not exist in the training set and are used to evaluate the generalization ability of existing approaches on unseen categories. We provide the detailed results in Section~\ref{sec:exp}. 

\begingroup
\setlength{\tabcolsep}{1.4pt}
\begin{table}[t]
\footnotesize
\centering
\caption{ Scale comparison between~\dataset~and existing datasets. ``Annotations'' denotes the total number of object annotations. ``Duration'' denotes the total duration (in minutes) of the annotated videos. }
\label{tab:dataset-cmp}
\vspace{5pt}
\begin{tabular}{|l|c|c|c|c|c|c|c|}
\hline
Scale & \begin{tabular}[x]{@{}c@{}}{JC}\\\cite{jumpcut}\end{tabular} & \begin{tabular}[x]{@{}c@{}}{ST}\\\cite{Fli2013segtrack}\end{tabular}   & \begin{tabular}[x]{@{}c@{}}{YTO}\\\cite{Jain2014youtubeobjects}\end{tabular} &  \begin{tabular}[x]{@{}c@{}}{FBMS}\\\cite{ochs2014FBMS}\end{tabular} & \multicolumn{2}{c|}{\begin{tabular}[x]{@{}c@{}}\multicolumn{1}{c}{DAVIS}\\\cite{Perazzi2016davis}~~\cite{Pont-Tuset2017davis}\end{tabular} }  & \begin{tabular}[x]{@{}c@{}}{\textbf{~\dataset}}\\(\textbf{Ours})\end{tabular}\\
\hline
Videos & 22 & 14 & 96 & 59 & 50 & 90 & \textbf{4,453} \\
\hline
Categories & 14 & 11 & 10 & 16 & - & - & \textbf{94} \\ 
\hline
Objects & 22 & 24 & 96 & 139 & 50 & 205 & \textbf{7,755} \\ 
\hline
Annotations & 6,331 & 1,475 & 1,692 & 1,465 & 3,440 & 13,543 & \textbf{197,272}  \\
\hline
Duration & 3.52 & 0.59 & 9.01 & 7.70 & 2.88 & 5.17 & \textbf{334.81}  \\
\hline
\end{tabular}
\end{table}
\vspace{-5pt}
\setlength{\tabcolsep}{1.4pt}
\endgroup

\section{Related work} \label{sec:related}

In the past decades, several datasets~\cite{Jain2014youtubeobjects,jumpcut,Fli2013segtrack,brox2010BMS,ochs2014FBMS,galasso2013VSB100} have been created for video object segmentation. All of them are in small scales which usually contain only dozens of videos. In addition, their video content is relatively simple (\eg no heavy occlusion, camera motion or illumination change) and sometimes the video resolution is low. Recently, a new dataset called DAVIS~\cite{Perazzi2016davis,Pont-Tuset2017davis} was published and has become the benchmark dataset in this area. Its 2016 version contains 50 videos with a single foreground object per video while the 2017 version has 90 videos with multiple objects per video. In comparison to previous datasets~\cite{Jain2014youtubeobjects,jumpcut,Fli2013segtrack,brox2010BMS,ochs2014FBMS,galasso2013VSB100}, DAVIS has both higher-quality of video resolutions and annotations. In addition, their video content is more complicated with multi-object interactions, camera motion, and occlusions.

Early methods~\cite{Jain2014youtubeobjects,Nagaraja2015video,Faktor2014voting,papazoglou2013fast,brox2010object} for video object segmentation often solve some spatial-temporal graph structures with hand-crafted energy terms, which are usually associated with features including appearance, boundary, motion and optical flows. Recently, deep-learning based methods were proposed due to its great success in image segmentation tasks~\cite{shelhamer2017fcn,Chen2016Deeplab}. Most of these methods~\cite{Caelles2017osvos,Perazzi2017masktrack,Cheng2017segflow,Jain_2017_CVPR,Yang2018osmn} build their models based on an image segmentation network and do not involve sequential modeling. Online learning~\cite{Caelles2017osvos} is commonly used to improve their performance. To make the model temporally consistent, the predicted mask of the previous frame is used as a guidance in~\cite{Perazzi2017masktrack,Yang2018osmn,Hu2017Maskrnn}. Other methods have been proposed to leverage spatial-temporal information. Jampani~\etal~\cite{Jampani2017vpn} use spatial-temporal consistency to propagate object masks over time. Tokmakov~\etal~\cite{Tokmakov2017memory} use a two-stream network to model objects' appearance and motion and use a recurrent layer to capture the evolution. However, due to the lack of training videos, they use a pretrained motion segmentation model~\cite{Tokmakov2017mpnet} and optical-flow model~\cite{Ilg2017flownet}, which leads to suboptimal results since the model is not trained end-to-end to best capture spatial-temporal features. Recently, Xu~\etal~\cite{xu2018youtube} propose a sequence-to-sequence learning algorithm to learn long-term spatial-temporal information for segmentation. Their models are trained on a preliminary version of ~\dataset~and do not depend on existing optical flow or motion segmentation models.

\section{\dataset}\label{sec:dataset}

\begingroup
\setlength{\tabcolsep}{1.4pt}
\begin{table}[t]
\footnotesize
\centering
\caption{ A complete list of object categories and number of instances in~\dataset. Objects are sorted from most frequent to least frequent.}
\label{tab:categories}
\vspace{5pt}
\begin{tabular}{|l|l|l|l|l|l|l|l|l|l|ll}
\hline
person       & 1702 & cat       & 115 & train         & 77 & hedgehog       & 49 & squirrel     & 24 & \multicolumn{1}{l|}{table}        & \multicolumn{1}{l|}{10} \\ \hline
ape          & 239  & snake     & 114 & owl           & 74 & eagle          & 45 & rope         & 24 & \multicolumn{1}{l|}{camera}       & \multicolumn{1}{l|}{10} \\ \hline
parrot       & 222  & zebra     & 111 & plant         & 73 & snail          & 44 & chameleon    & 22 & \multicolumn{1}{l|}{watch}        & \multicolumn{1}{l|}{9}  \\ \hline
giant\_panda & 222  & giraffe   & 110 & airplane      & 73 & toilet         & 43 & box          & 20 & \multicolumn{1}{l|}{stuffed\_toy} & \multicolumn{1}{l|}{9}  \\ \hline
sedan        & 221  & bear      & 97  & bus           & 70 & camel          & 40 & tissue       & 18 & \multicolumn{1}{l|}{guitar}       & \multicolumn{1}{l|}{8}  \\ \hline
lizard       & 189  & fox       & 90  & shark         & 66 & frisbee        & 39 & kangaroo     & 18 & \multicolumn{1}{l|}{microphone}   & \multicolumn{1}{l|}{7}  \\ \hline
duck         & 186  & leopard   & 88  & tiger         & 66 & whale          & 38 & cloth        & 18 & \multicolumn{1}{l|}{cup}          & \multicolumn{1}{l|}{6}  \\ \hline
dog          & 177  & elephant  & 87  & surfboard     & 64 & knife          & 38 & bottle       & 17 & \multicolumn{1}{l|}{shovel}       & \multicolumn{1}{l|}{6}  \\ \hline
skateboard   & 173  & horse     & 87  & earless\_seal & 63 & tennis\_racket & 38 & small\_panda & 16 & \multicolumn{1}{l|}{flag}         & \multicolumn{1}{l|}{6}  \\ \hline
monkey       & 164  & others    & 86  & frog          & 63 & crocodile      & 37 & spider       & 14 & \multicolumn{1}{l|}{mirror}       & \multicolumn{1}{l|}{5}  \\ \hline
sheep        & 155  & deer      & 86  & mouse         & 63 & umbrella       & 36 & ball         & 14 & \multicolumn{1}{l|}{ring}         & \multicolumn{1}{l|}{5}  \\ \hline
fish         & 138  & motorbike & 85  & boat          & 61 & paddle         & 33 & jellyfish    & 13 & \multicolumn{1}{l|}{necklace}     & \multicolumn{1}{l|}{4}  \\ \hline
rabbit       & 135  & turtle    & 84  & snowboard     & 59 & raccoon        & 29 & eyeglasses   & 11 & \multicolumn{1}{l|}{ant}          & \multicolumn{1}{l|}{3}  \\ \hline
hat          & 131  & bird      & 81  & penguin       & 53 & parachute      & 28 & backpack     & 11 &                                   &                         \\ \cline{1-10}
cow          & 128  & truck     & 81  & lion          & 52 & bucket         & 28 & butterfly    & 11 &                                   &                         \\ \cline{1-10}
hand         & 121  & dolphin   & 80  & sign          & 50 & bike           & 28 & handbag      & 11 &                                   &                         \\ \cline{1-10}
\end{tabular}
\end{table}
\vspace{-5pt}
\setlength{\tabcolsep}{1.4pt}
\endgroup

To create our dataset, we first carefully select a set of video categories including animals (\eg \textit{ant, eagle, goldfish, person}), vehicles (\eg \textit{airplane, bicycle, boat, sedan}), accessories (\eg \textit{eyeglass, hat, bag}), common objects (\eg \textit{potted plant, knife, sign, umbrella}), and humans in various activities (\eg \textit{tennis, skateboarding, motorcycling, surfing}). The videos containing human activities have diversified appearance and motion, so we collect human-related videos using a list of activity tags to increase the diversity of human motion and behaviors. Most of these videos contain interactions between a person and a corresponding object, such as tennis racket, skateboard, motorcycle, etc. The entire category set includes 78 categories that covers diverse objects and motions, and should be representative for everyday scenarios.



We then collect many high-resolution videos with the selected category labels from the large-scale video classification dataset YouTube-8M~\cite{abu2016youtube}. This dataset consists of millions of YouTube videos associated with more than 4,700 visual entities. We utilize its category annotations to retrieve candidate videos that we are interested in. Specifically, up to $100$ videos are retrieved for each category in our segmentation category set. There are several advantages to using YouTube videos to create our segmentation dataset. First, YouTube videos have very diverse object appearances and motions. Challenging cases for video object segmentation, such as occlusions, fast object motions and change of appearances, commonly exist in YouTube videos. Second, YouTube videos are taken by both professionals and amateurs and thus different levels of camera motions are shown in the crawled videos. Algorithms trained on such data could potentially handle camera motion better and thus are more practical. Last but not the least, many YouTube videos are taken by today's smart phone devices and there are demanding needs to segment objects in those videos for applications such as video editing and augmented reality. 

\begingroup
\setlength{\tabcolsep}{1pt}
\renewcommand{\arraystretch}{1}
\begin{figure*}[t]
\centering
 \begin{tabular}{@{}cccc@{}}
\includegraphics[width=.21\textwidth]{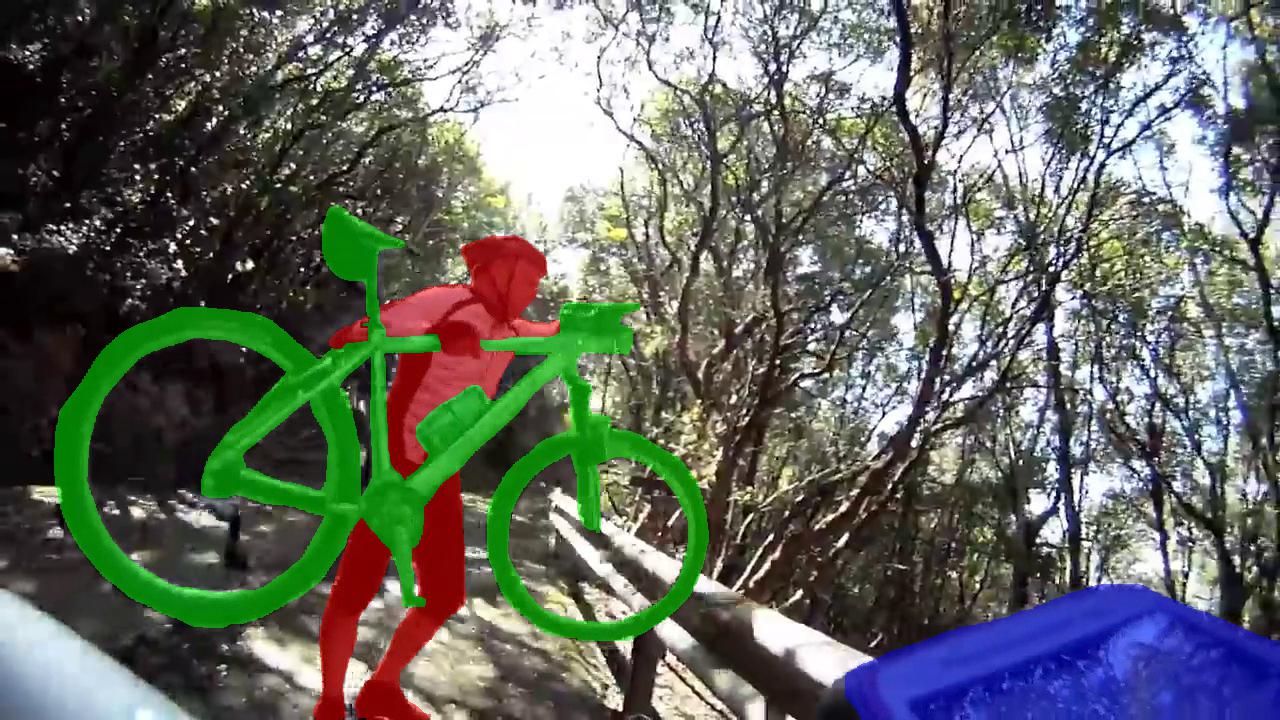}  &
\includegraphics[width=.21\textwidth]{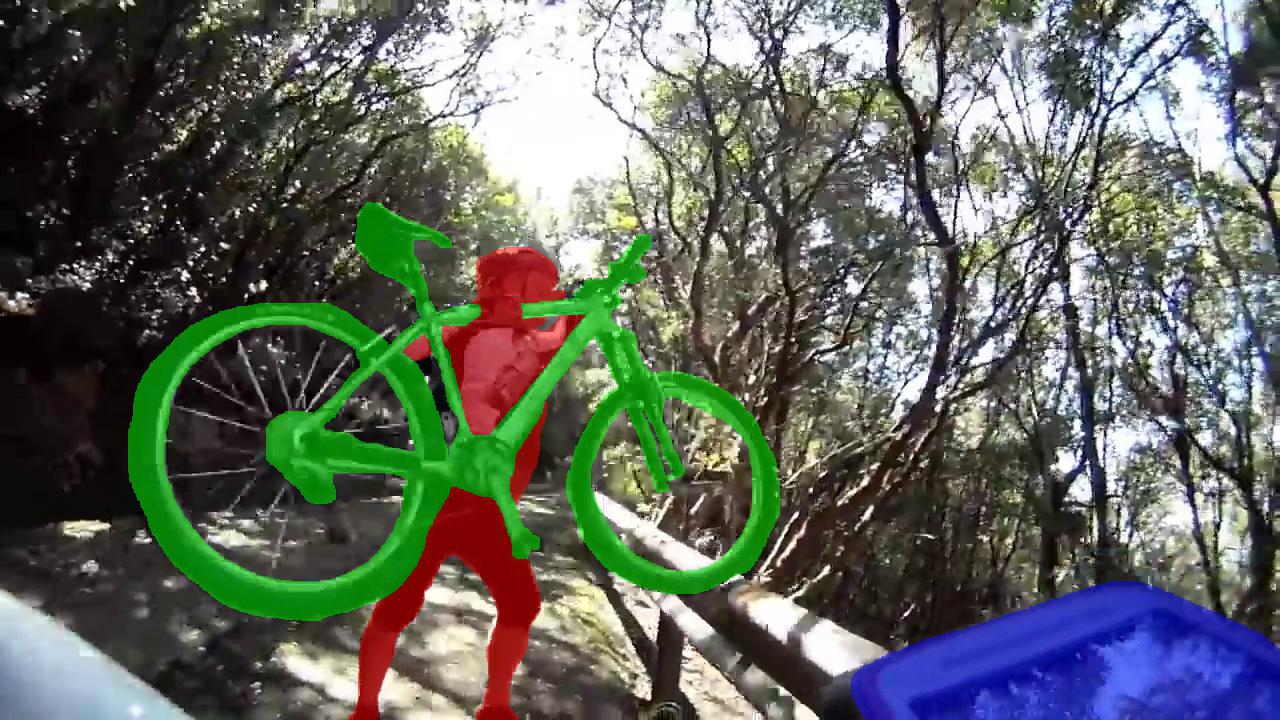}  &
 \includegraphics[width=.21\textwidth]{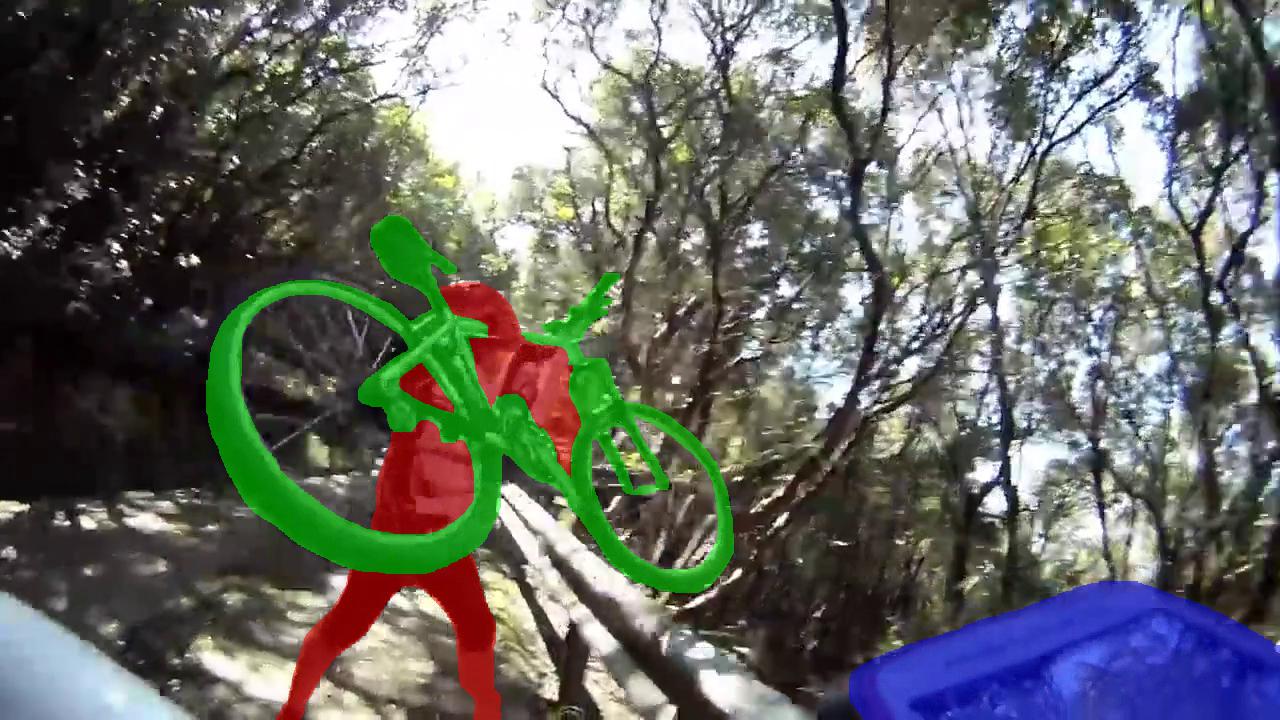}  &
 \includegraphics[width=.21\textwidth]{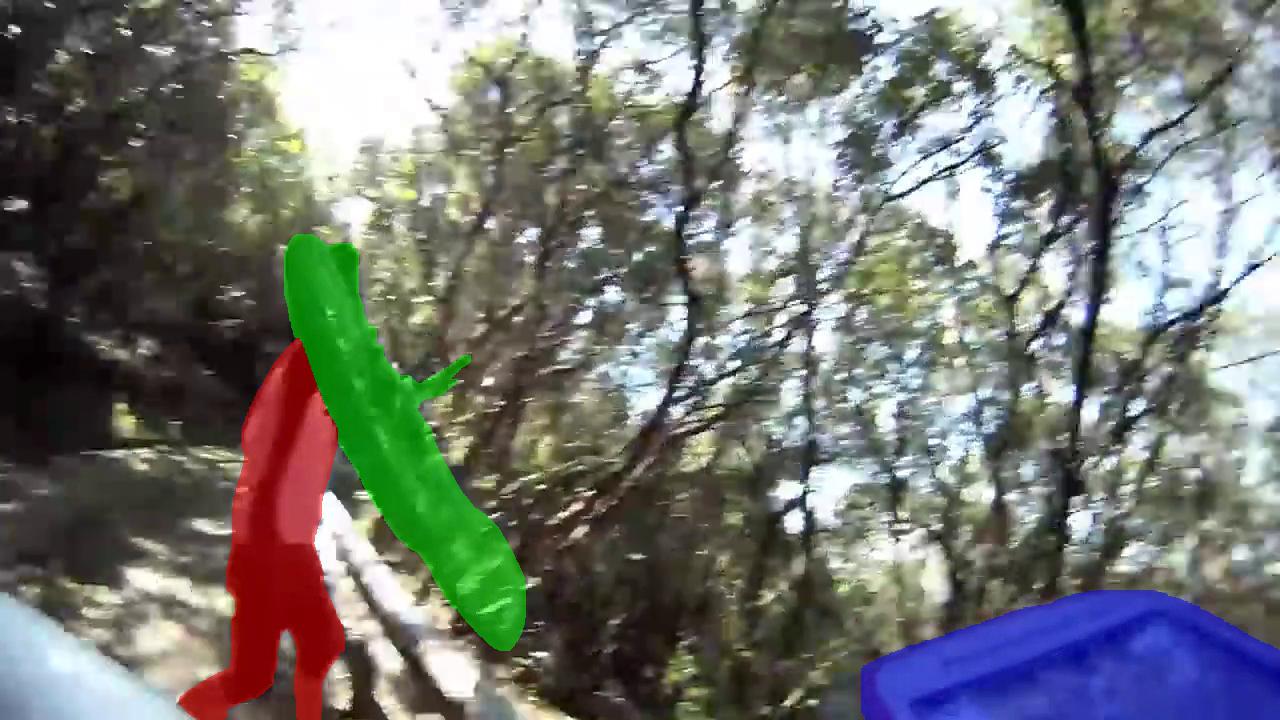}  \\
 \includegraphics[width=.21\textwidth]{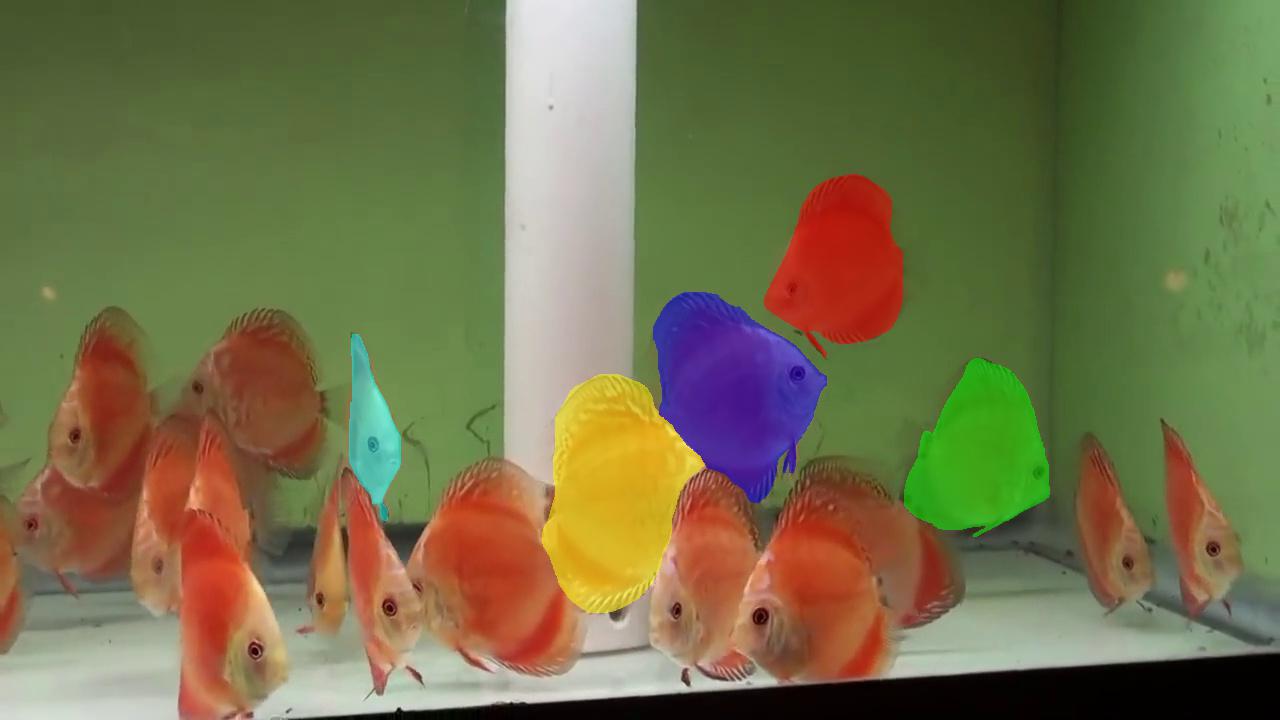}  &
\includegraphics[width=.21\textwidth]{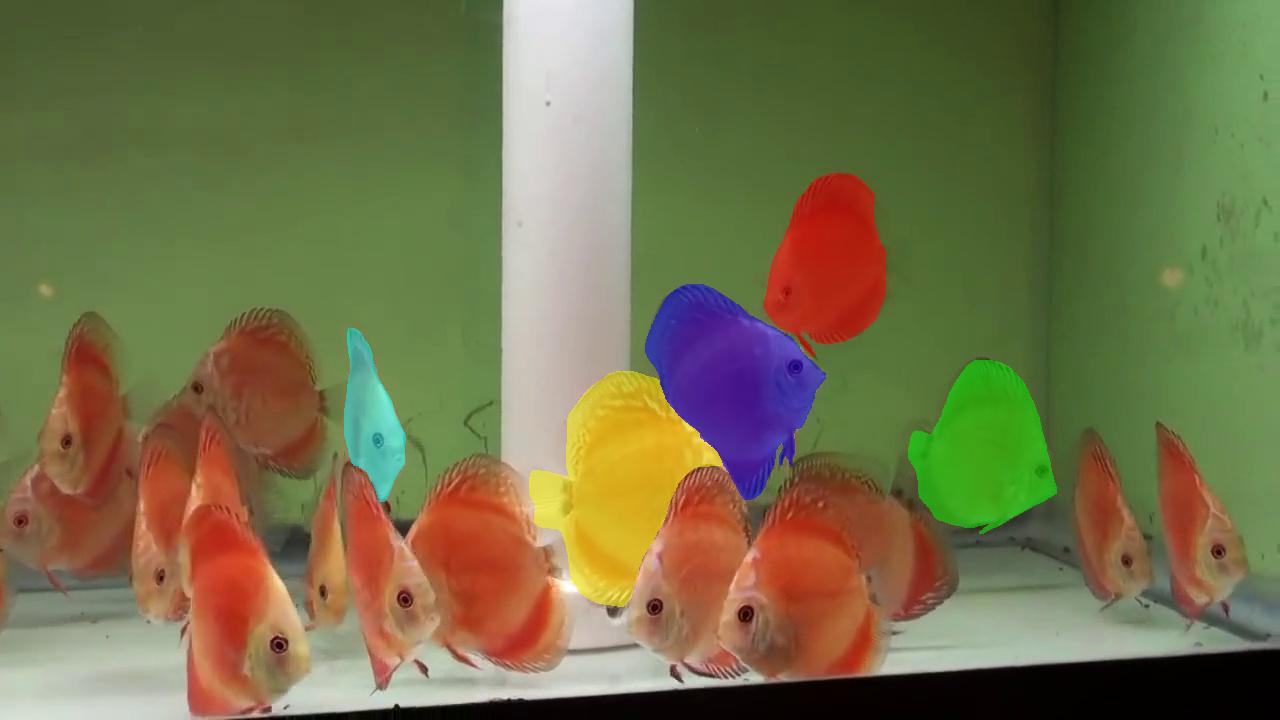}  &
 \includegraphics[width=.21\textwidth]{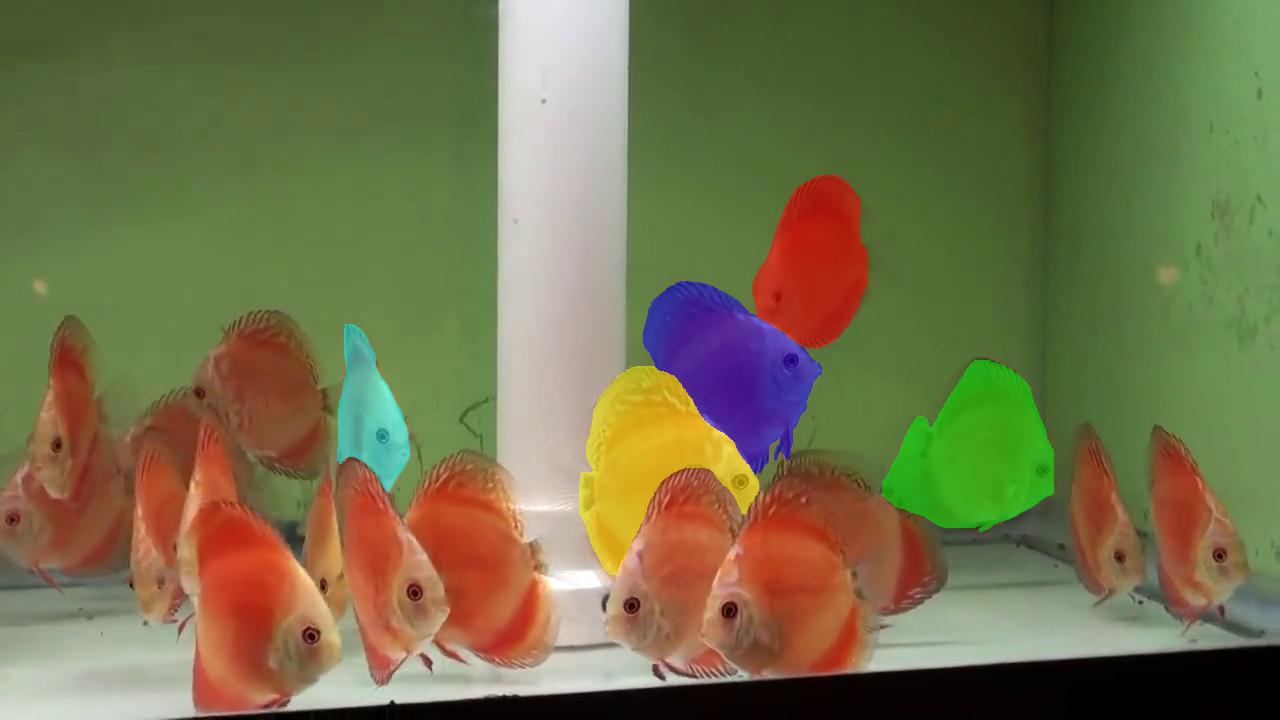}  &
 \includegraphics[width=.21\textwidth]{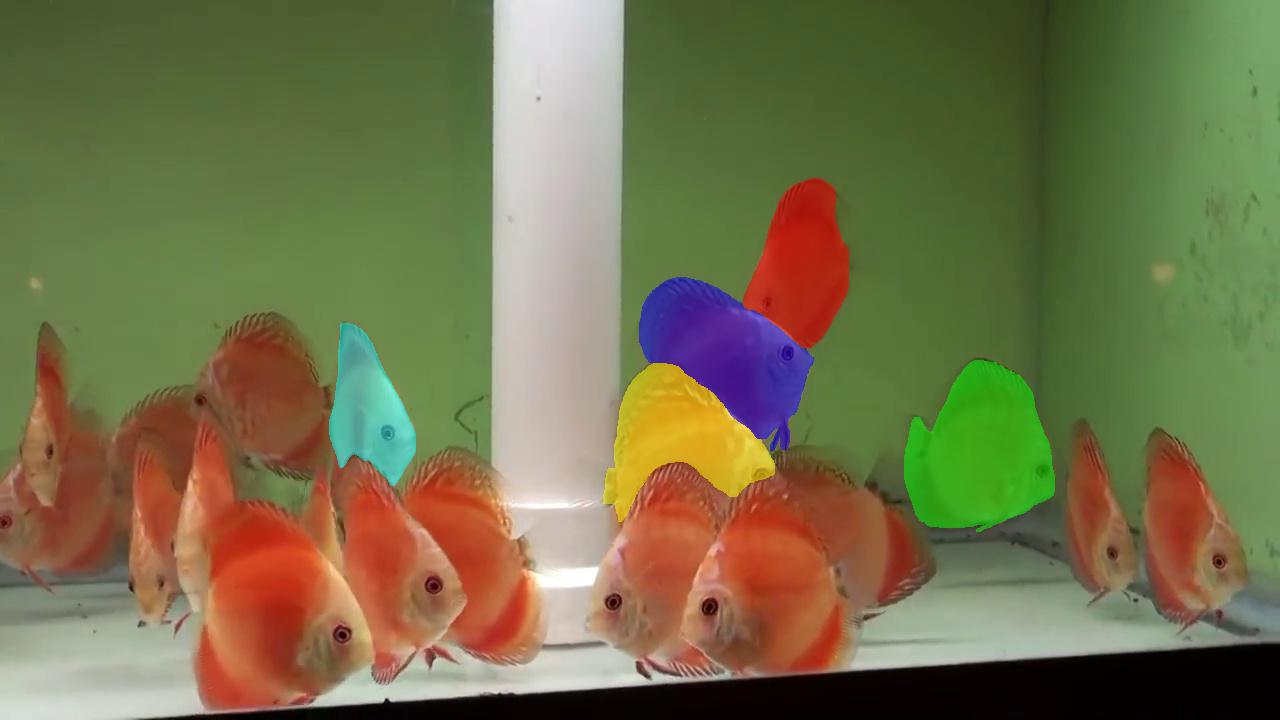}  
 \end{tabular}
\vspace{-10pt}
 \caption{The ground truth annotations of sample video clips in our dataset. Different objects are highlighted with different colors.}
\label{fig:dataset_example}
\end{figure*}
\setlength{\tabcolsep}{1.4pt}
\endgroup

Since the retrieved videos are usually long (several minutes) and have shot transitions, we use an off-the-shelf video shot detection algorithm~\footnote{http://johmathe.name/shotdetect.html} to automatically partition each video into multiple video clips. We first remove the clips from the first and last 10\% of the video, since these clips have a high chance of containing introductory subtitles and credits lists. We then sample up to five clips with appropriate lengths (3$\sim$6 seconds) per video and manually verify that these clips contain the correct object categories and are useful for our task (\eg no scene transition, not too dark, shaky, or blurry).
After the video clips are collected, we ask human annotators to select up to five objects of proper sizes and categories per video clip and carefully annotate them (by tracing their boundaries instead of rough polygons) every five frames in a $30$fps frame rate, which results in a $6$fps sampling rate. Given a video and its category, annotators are first required to annotate objects belonging to that category. If the video contains other salient objects, we ask the annotators to label them as well, so that each video has multiple objects annotated, and the object categories are not limited to our initial 78 categories. In human activity videos, both the human subject and the object he/she interacts with are labeled, \eg, both the person and the skateboard are required to be labeled in a ``skateboarding'' video. Further, the instance-level categories are labeled for each annotated object, including not only the video-level categories, but also additional categories that the labelers has labeled, resulting in a total of 94 object categories. The activity categories are removed since they do not represent single objects. Note in an earlier version of the dataset~\cite{xu2018youtube}, only video-level categories are available. Some annotation examples are shown in Figure~\ref{fig:dataset_example}. Unlike dense per-frame annotation in previous datasets~\cite{jumpcut,Perazzi2016davis,Pont-Tuset2017davis}, we believe that the temporal correlation between five consecutive frames is sufficiently strong that annotations can be omitted for intermediate frames to reduce the annotation efforts. Such a skip-frame annotation strategy allows us to scale up the number of videos and objects under the same annotation budget, which are important factors for better performance. We find empirically that our dataset is effective in training different segmentation algorithm.


As a result, our collected dataset~\dataset~consists of 4,453 YouTube video clips which is about 50 times larger than YouTubeObjects~\cite{Jain2014youtubeobjects}, the existing video object segmentation dataset with the most videos. Our dataset also has a total of 197,272 object annotations which is 15 times larger than those of DAVIS 2017~\cite{Pont-Tuset2017davis}. There are 94 different object categories including person, animals, vehicles, furnitures, and other common objects. A complete list of object categories can be seen in Table~\ref{tab:categories}. Therefore,~\dataset~is the largest and most comprehensive dataset for video object segmentation to date. 

\section{Experiments} \label{sec:exp}

In this section, we retrain state-of-the-art video object segmentation methods on~\dataset~training set and evaluate their performance on~\dataset~validation set. All the algorithms are trained and tested under the same setting. We hope the experiment results could setup baselines for the development of new algorithms in the future.

\subsection{Settings}

The whole dataset which consists of 4,453 videos is split into training (3,471), validation (474) and test (508) sets. Since the dataset has been used for a workshop competition (\ie The 1st Large-scale Video Object Segmentation Challenge)~\footnote{\href{https://youtube-vos.org/challenge/challenge2018}{https://youtube-vos.org/challenge/challenge2018}}, the test set will only be available during the competition period while the validation set will be always publicly available. Therefore we only use the validation set for evaluation. In the training set, there are 65 unique object categories which are regarded as seen categories. In the validation set, there are 91 unique object categories which include all the seen categories and 26 unseen categories. As stated, the unseen categories are used to evaluate the generalization ability of different algorithms. For training the state-of-the-arts algorithms, we first resize the training frames to a fixed size (\ie 256$\times$448) and then use their publicly released codes to train their models. We also evaluate the algorithms on other image resolutions such as $480$p but the difference is negligible. All the models are trained sufficiently until convergence. For evaluation, we follow the evaluation method used by the workshop, which computes the region similarity \(\mathcal{J}\) and the contour accuracy \(\mathcal{F}\) as in~\cite{Perazzi2016davis}. The final result is the average of four metrics: \(\mathcal{J}\) for seen categories, \(\mathcal{F}\) for seen categories, \(\mathcal{J}\) for unseen categories, and \(\mathcal{F}\) for unseen categories. 

\subsection{Methods}

We compare several recently proposed algorithms which achieved state-of-the-art results on previous small-scale benchmarks. These algorithms are {OSVOS}~\cite{Caelles2017osvos}, {MaskTrack}~\cite{Perazzi2017masktrack}, {OSMN}~\cite{Yang2018osmn}, {OnAVOS}~\cite{voigtlaender2017online} and {S2S}~\cite{xu2018youtube}. For more details of these algorithms, please refer to their papers.

\subsection{Results}

The results are presented in Table~\ref{tab:res_256}. The first four methods use static image segmentation models and three of them (\ie OSVOS, MaskTrack and OnAVOS) require online learning. S2S leverages long-term spatial-temporal coherence by recurrent neural networks (RNN) and its model without online learning (the second last row in Table~\ref{tab:res_256}) achieves comparable performance compared to the best results of online-learning methods, which effectively demonstrates the importance of long-term spatial-temporal information for video object segmentation. With online learning, S2S is further improved and achieves around $6\%$ absolute improvement over the best online-learning method OSVOS on overall accuracy. Surprisingly, OnAVOS which is the best performing method on DAVIS does not achieve good results on our dataset. We believe the drastic appearance changes and complex motion patterns in our dataset makes the online adaptation fail in many cases.

Next we compare the generalization ability of existing methods on unseen categories in Table~\ref{tab:res_256}. All the methods have obviously better results on seen categories than unseen categories. Among them, OSVOS has the least discrepancy, possibly due to the pre-training on large-scale image segmentation dataset. It is also worth noting that methods with online learning also suffer from this problem, which suggests that although online learning is helpful to improve the accuracy on unseen categories, pre-training on some large-scale object segmentation dataset is still important to learn general object feature representation. In general, the results shows a much larger performance gap between seen and unseen categories compared to~\cite{xu2018youtube}. We believe it is because instance categories are used to split the seen and unseen subset in the current setting, comparing to~\cite{xu2018youtube} in which subsets are split using video-level categories. The current setting leads to a clearer separation between seen and unseen categories and is more challenging.

Lastly we compare the inference speed of all the methods averaged per frame. OSMN and S2S (w/o OL) do not use online learning and thus have very fast inference speed, which can be applied in real time. This is a big advantage over those online learning methods especially for mobile applications. While the performance is still inferior to online learning ones. 


\begin{table*}[t]
\centering
\caption{Comparisons of state-of-the-art methods on~\dataset~validation set. ``$\mathcal{J}$" and ``$\mathcal{F}$" denote the region similarity and the contour accuracy. ``seen'' and ``unseen'' denote the results averaged over the seen categories and unseen categories. ``Overall'' denote the  results averaged over the four metrics. ``OL" denotes online learning. The best results are highlighted in bold.}
\label{tab:res_256}
\begin{tabular}{|l|c|c|c|c|c|c|}
\hline
Method       & $\mathcal{J}$ seen & $\mathcal{J}$ unseen & $\mathcal{F}$ seen & $\mathcal{F}$ unseen  & Overall & \makecell{Speed \\(s/frame)} \\ \hline
OSVOS~\cite{Caelles2017osvos}  & 59.8\%	& 54.2\%&	60.5\% &	60.7\% &	58.8\% & 10   \\ \hline
MaskTrack~\cite{Perazzi2017masktrack} & 59.9\%   & 45.0\% & 59.5\%  & 47.9\%  &  53.1\% & 12  \\ \hline
OSMN~\cite{Yang2018osmn} & 60.0\%	& 40.6\%	 & 60.1\%	& 44.0\%	& 51.2\% & \textbf{0.14}       \\ \hline
OnAVOS~\cite{voigtlaender2017online} & 60.1\%  & 46.6\%   & 62.7\%  & 51.4\% & 55.2\%  & 13    \\ \hline
S2S (w/o OL)~\cite{xu2018youtube}        &  66.7\%  &  48.2\%   & 65.5\% & 50.3\%  &  57.6\%  & 0.16  \\ \hline
S2S (with OL)~\cite{xu2018youtube}  & \textbf{71.0\%}     & \textbf{55.5\%}   &  \textbf{70.0\%}   & \textbf{61.2\%}    & \textbf{64.4\%}   & 9   \\ \hline
\end{tabular}
\end{table*}

\section{Conclusion} \label{sec:conclusion}
In this report, we introduce the largest video object segmentation dataset to date. The new dataset called \dataset, much larger than existing datasets in terms of number of videos and annotations, allows us to evaluate existing state-of-the-art video object segmentation methods more comprehensively. We believe the new dataset will foster research on video-based computer vision in general.


\bibliographystyle{splncs}
\bibliography{egbib}
\end{document}